\documentclass[conference]{IEEEtran}

\usepackage{amsmath,amssymb,amsthm,amsfonts,enumitem}
\usepackage{algorithm}
\usepackage{algpseudocode}
\usepackage{graphicx}
\usepackage{textcomp}
\usepackage{xcolor}
\usepackage{xspace}
\usepackage{subcaption}
\usepackage{multirow}
\usepackage{xcolor}
\usepackage{tikz}
\usepackage[hidelinks]{hyperref}
\usepackage[numbers,sort&compress]{natbib}

\usepackage{cleveref}
\crefname{section}{\S}{\S\S}

\usepackage{enumitem}
\setlist[itemize]{left=0pt}

\usepackage{listings}
\definecolor{datatypecolor}{rgb}{0.75, 0.0, 0.2}
\definecolor{keywordcolor}{rgb}{0.01, 0.28, 1.0}
\def\BibTeX{{\rm B\kern-.05em{\sc i\kern-.025em b}\kern-.08em
    T\kern-.1667em\lower.7ex\hbox{E}\kern-.125emX}}

\newcommand{\projname}{DISTWAR\xspace}
\newcommand{\SWprojname}{DISTWAR\xspace}

\newcommand*\circled[1]{\tikz[baseline=(char.base)]{
            \node[shape=circle,fill,inner sep=1pt] (char) {\textcolor{white}{#1}};}}

\title{\huge\textbf{\textit{\projname}}: Fast Differentiable Rendering\\ on Raster-based Rendering Pipelines}

\author{
  \IEEEauthorblockN{
    Sankeerth Durvasula\IEEEauthorrefmark{1},     
    Adrian Zhao\IEEEauthorrefmark{1},     
    Fan Chen,     
    Ruofan Liang,     
    Pawan Kumar Sanjaya,     
    Nandita Vijaykumar\\
  }
  \IEEEauthorblockA{
    University of Toronto\\
    \texttt{\{sankeerth,adrianz,fan,ruofan,pawan,nandita\}@cs.toronto.edu}
  }
}

\begin{document}
\maketitle
\thispagestyle{plain}
\pagestyle{plain}


\begin{abstract}
Differentiable rendering is a technique used in an important emerging class of visual computing applications that involves representing any 3D scene as a model that is trained from 2D images using gradient descent.
Recent works (e.g., 3D Gaussian Splatting) integrate the rasterization pipeline to enable rendering high quality photo-realistic imagery at high speeds from these learned 3D models. 
These methods have been demonstrated to be very promising, providing state-of-art quality for many important tasks. However, training a model to represent a scene is still a time-consuming task even when using powerful GPUs. 
In this work, we observe that the gradient computation phase during training is a significant bottleneck on GPUs due to the large number of \emph{atomic operations} that need to be processed. These atomic operations overwhelm the atomic units in the L2 subpartitions causing long stalls. 

To address this challenge, we leverage the observations that during the gradient computation: (1) for most warps, all threads atomically update the same memory locations; and (2) warps generate varying amounts of atomic traffic (since some threads may be inactive). We propose \projname, a primitive that accelerate atomic operations based on two key ideas: First, we enable warp-level reduction of threads at the SM sub-cores using registers to leverage the locality in intra-warp atomic updates. Second, we distribute the atomic computation between the warp-level reduction at the SM and the L2 atomic units to increase the throughput of atomic computation. Warps with many threads performing atomic updates to the same memory locations are scheduled at the SM, and the rest using existing L2 atomic units. 
We propose a software-only implementations of \projname that using existing warp-level primitives.
We evaluate \projname on real GPUs using widely used raster-based differentiable rendering workloads. We demonstrate significant speedups of 2.44× on average (and up to 5.7×).

\end{abstract}

\section{Introduction}

Differentiable rendering leverages machine learning to solve some fundamental tasks in computer graphics, such as scene reconstruction~\citep{nerf, 3dgs} (deriving a representation of a 3D scene),
and inverse rendering~\cite{mitsuba, nvdiffrec} (estimating shape, texture, lighting, and material of a 3D object) from a set of rendered/captured reference images.
These problems are central to many important applications~\cite{tewari2020state}, such as photogrammetry, 3D modeling and scanning, 3D model creation tools, game engines, and AR/VR applications. 
With differentiable rendering, these tasks are formulated as a learning problem that can be solved using gradient descent-based optimization techniques. 

For example, 
neural radiance fields (NeRFs)~\cite{nerf, instantngp, mipnerf, tancik2023nerfstudio, mipnerf360, zipnerf,tensorf, fastnerf, bakingnerf, nsvf, plenoctrees, donerf, kilonerf, dvgo, plenoxels}
is a popular and promising approach to capture high quality photo-realistic representations of the environment. They represent a scene using a set of learnable parameters (i.e., a model, typically structured as a 3D grid along with a neural network). Any 2D view of the scene can then be rendered using this representation. These model parameters are trained to represent a scene using gradient descent by computing a loss function between a ground truth image and the image generated by rendering the current model. A differentiable renderer is used to compute the loss gradients with respect to these scene parameters.  
Learning-based methods for these tasks have demonstrated significant success in achieving state-of-art accuracy in scene reconstruction, leading to huge interest in the computer graphics, vision, and robotics communities. This success has led to the development of several specialized frameworks and libraries for differentiable rendering~\cite{nvdiffrast, pytorch3d, mitsuba, drjit}, and recent work~\cite{slang, slangd} propose native support for differentiable rendering in GPUs as a first-class feature. Prior works~\cite{li2022rt, neurex, instant3disca, artist,  mubarik2023hardware} have also proposed accelerators for NeRF-based rendering and training. 

Another more recent approach for differentiable rendering is to leverage the high-speed rasterization pipeline~\cite{opengl} in GPUs. Rasterization requires the scene to be represented as a set of \emph{geometric primitives} (i.e. meshes, triangles, points) in 3D space which can then be rendered as 2D images with very high speeds. 
Differentiable rendering with rasterization involves learning these primitives using similar gradient descent-based training. This approach~\cite{pulsar, nvdiffrec, adop, 3dgs} 
has demonstrated state-of-art capability in producing high-quality scene reconstructions at high speeds, and has emerged as a promising representation for 3D visual data. 
Among these methods, a recent transformative work, is 3D Gaussian Splatting (3DGS)~\cite{3dgs} and has spurred significant interest in both industry and academia~\cite{3dgs_avatar, dynamic3dgs, deform3dgs, 3dgsdreamer, 3dgs_flexiblerendering, 3dgs_markerless_motioncapture, 4dgs, gaussianrep_survey}.
3DGS represents the scene geometry with 3D Gaussians as its primitives (that are associated with learnable parameters) 
and uses an efficient tile-based rasterizer~\cite{adop, pulsar} to render images from the Gaussians.  

While rendering scene representations with learnable parameters can be done at high speeds using the raster-based rendering pipeline, \emph{training} these models to learn scenes can still be a slow process requiring many hours for each scene on a powerful GPU. In this work, we perform a detailed performance analysis of differentiable rendering applications.  
We find that the gradient computation step of the backward pass (which involves computing and aggregating gradients with respect to trainable scene parameters) is a significant bottleneck. For example, in \texttt{3DGS} workloads, the gradient computation takes up on average 30.07\% (up to 65.8\%) of the overall training time on the RTX 4090 GPU (\cref{sec:motivation_backward_runtime}).

Our analysis shows that this bottleneck is primarily caused by a large number of atomic operations that accumulate gradients for the model parameters. During the gradient computation, each thread is associated with one pixel. 
These gradient updates must be done using atomic operations since multiple threads may update the same set of parameters.
Since each thread updates many parameters, this leads to a massive number of atomic operations.
These atomic operations cause significant contention at the atomic units at the L2 memory subpartitions (ROP units), leading to long stalls at the GPU streaming multiprocessors (SMs) (\cref{sec:motivation_shaderbwdimpl}).


Our \textbf{goal} in this work is to accelerate raster-based differentiable rendering applications by accelerating atomic operations that constitute a significant bottleneck during the gradient computation. 
From our analysis of the atomic operations in gradient computation, we make two observations: \textbf{(1) Locality in intra-warp atomic updates:} Threads \emph{within a warp} typically update the same parameters and thus the \emph{same memory location}. For example, for the \texttt{3D-PR} workload, we find that over 99\% of warps have all its threads update the same memory location  (\cref{sec:motivation_observations}).   
\textbf{(2) Only a subset of threads in a warp perform atomic updates:} There is significant variation in the number of threads within each warp that make gradient updates at any time (\cref{sec:motivation_observations}) as some threads are made inactive due to failing condition checks in the code (i.e., control divergence). The number of threads making atomic requests determines the atomic request traffic generated by the warp and varies across warps.

Prior approaches~\cite{lab, phi, hlrc} that address bottlenecks due to atomic requests in GPUs, buffer and aggregate atomic updates in the L1 cache to reduce traffic in the interconnect and L2 atomic units (ROP units). 
While these approaches can effectively alleviate overheads from atomic operations for a wide range of applications, they do not leverage the intra-warp locality in atomic updates seen in differentiable rendering. The sheer number of atomic requests generated also overwhelm the load-store units before the atomics can be aggregated, making this approach less effective for differentiable rendering workloads (\cref{sec:related_work}). 


In this work, we introduce \projname (\underline{Dist}ributed-\underline{W}arp-level and \underline{A}tomic-Unit collaborative \underline{R}eduction), a primitive that accelerates atomic
updates in applications that (1) generate significantly large amounts of
atomic requests and (2) typically have
most threads within an warp performing atomic updates to
the same memory locations. \projname is based on two key ideas: \textbf{(1)} We leverage intra-warp locality in atomic updates (Observation 1) to perform \emph{warp-level reduction} at the core itself using registers. This significantly reduces the number of atomic operations that need to be sent to the L2 atomic units to update global memory. \textbf{(2)} We dynamically distribute the atomic computation between the cores and L2 atomic units to enable high throughput atomic updates by leveraging all atomic units. Leveraging Observation 2, warps that only generate a few atomic updates are handled at the L2 atomic units. Warps where most/all threads generate atomic updates are first reduced at the SM using the proposed warp-level reduction.
Implementing \projname requires addressing important design challenges (described in~\cref{sec:design_challenges}). We propose a software-only implementation of \projname that leverages existing warp-level  primitives (such as \texttt{\_\_shfl\_sync}) to implement warp-level reduction at each SM sub-core. Atomic updates to any memory location involving more than a predefined number of threads in a warp are performed at the SM, and the rest is performed at the ROP units. This predefined number is a tunable hyperparameter (the \textit{balancing threshold}).

We evaluate \projname across recent widely used differentiable rendering applications (3D Gaussian Splatting~\cite{3dgs}, NVDiffRec~\cite{nvdiffrec}, Pulsar~\cite{pulsar, pytorch3d}). With \projname, we demonstrate a speedup of $2.6\times$ on average (up to $5.7\times$) for gradient computation and an average speedup of $1.41\times$ (up to $2.4\times$) on the overall application on a real NVIDIA RTX 4090 GPU. Our contributions are summarized as follows: 
\begin{itemize}
\item This is the first work to perform a performance characterization of an important emerging workload, rasterization-based differential rendering for 3D visual data, and identify atomic updates as a key bottleneck.
\item We introduce \projname, a novel primitive to accelerate atomic processing in GPUs for applications that produce large amounts of atomic requests and with intra-warp locality in atomic updates. 
\item We will open-source \projname, which can be directly used to obtain significant speedups on raster-based differentiable rendering workloads.
\item We evaluate \projname on popular differentiable rendering applications on real hardware and demonstrate significant speedups.
\end{itemize}


\section{Background}
\noindent

\subsection{Atomic Processing in GPUs}
\label{sec:bckg_atomicprocessing}
\begin{figure}[!htb]
    \centering
    \vspace{-.2cm}
    \includegraphics[width=0.92\linewidth]{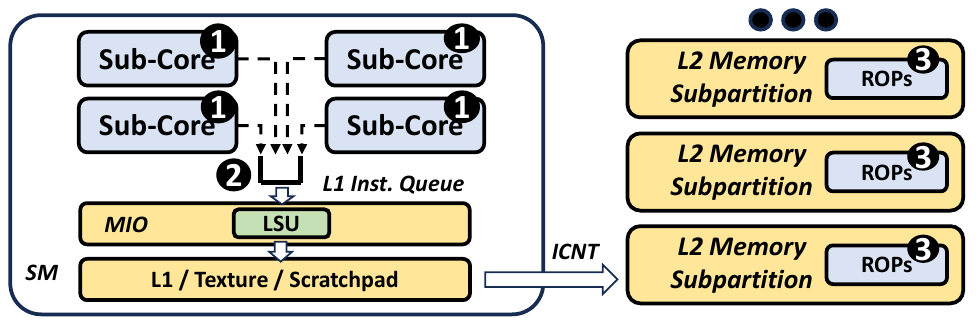}
    \caption{Atomic processing in a GPU.}
    \label{fig:gpu_baseline}
    \vspace{-.71cm}
\end{figure}

Fig.~\ref{fig:gpu_baseline} depicts a Streaming Multiprocessor (SM) of a modern GPU~\cite{voltaarch}. Each SM consists of multiple (typically $4$) sub-cores~\circled{1}. Each sub-core consists of its own warp scheduler, register file, and execution units. Each sub-core sends local, global and atomic memory requests to the MIO (Memory I/O Unit) which interfaces with the caches and memory subsystem through a queue~\cite{nsight_prof} (sometimes called L1 instruction queue \circled{2}). In this work, we refer to the unit that dispatches requests from the sub-cores to the caches and memory subsystem as the Load-Store Unit (LSU) (consistent with NVIDIA's NSight terminology~\cite{nsight_prof}). Atomic operations sent to the LSU are issued to the memory subpartition~\circled{3} via the interconnect. The memory subpartition contains compute units (known as ROP units)~\cite{amperearch, voltaarch, gpgpu_book} which process the atomic requests at the L2 caches which are shared across all SMs~\cite{rmo, t3ermo}.
A large number of atomic requests may lead to traffic in the interconnect and contention at the ROP units.

\subsection{Differentiable Rendering for 3D Scene Reconstruction}

\begin{figure}[!htb]
    \centering
        \vspace{-.30cm}

    \includegraphics[width=\linewidth]{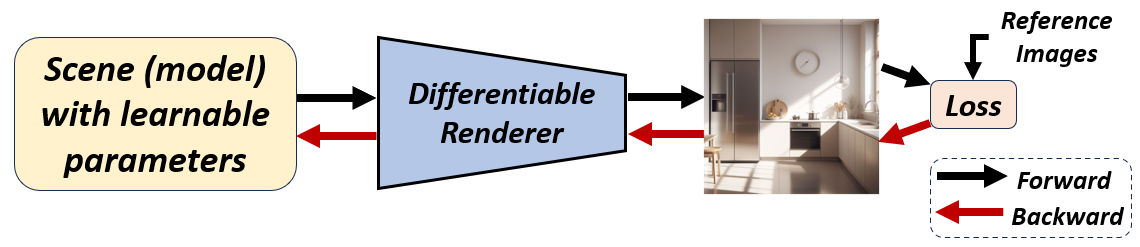}
    \vspace{-.5cm}
    \caption{A generalized differentiable rendering training pipeline to train a \emph{model} to learn a 3D scene.}
    \label{fig:ml_diffrendering}
    \vspace{-.2cm}
\end{figure}

We describe differentiable rendering using a classic and important problem in computer graphics: 3D scene reconstruction, which involves creating a 3D representation of a scene from 2D images. 
3D scene reconstruction has several important applications in novel view synthesis, 3D scanning and modelling, and photogrammetry. 
With differentiable rendering, the scene is represented using a set of parameters (i.e., model) that are learned using gradient descent, similar to standard deep learning training. This process of training a model to represent a 3D scene is depicted in Fig.~\ref{fig:ml_diffrendering}. 

An initialized model is rendered from a view point to generate a 2D image (i.e., the forward pass in Fig.~\ref{fig:ml_diffrendering}). The difference between the rendered image and the corresponding reference/ground truth image (i.e., \emph{loss}) is obtained by subtracting their RGB values. This loss is backpropagated to calculate gradients for all model parameters that minimize the loss using gradient descent-based optimization (the backward pass in Fig.~\ref{fig:ml_diffrendering}). This process is repeated for images from different view points. Examples of such models, also referred to as \emph{implicit representations}, include neural radiance fields (NeRFs)~\cite{nerf} and 3D Gaussians~\cite{3dgs}. These approaches have been transformative in representing visual data (e.g., 3D scenes, images, and videos), generating significant interest in industry and academia, due to the differentiability and compactness of the representation and the state-of-art performance in novel-view synthesis. 

\subsection{Differentiable Rendering for Rasterization Pipelines}
\label{sec:bckg_diffrendering}
\begin{figure}[!htb]
    \centering
    \includegraphics[width=0.98\linewidth]{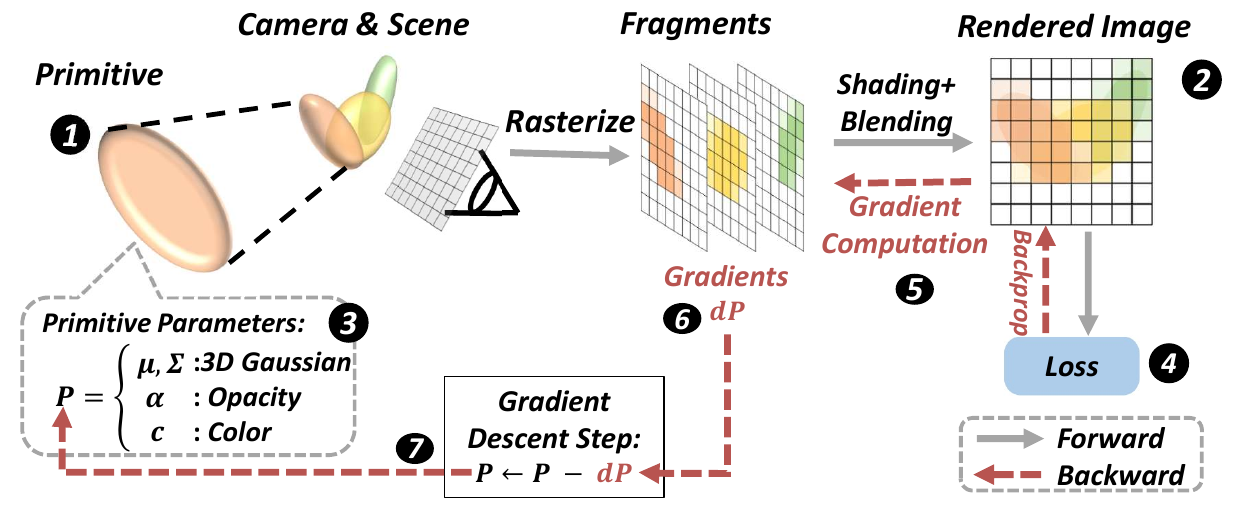}
    \caption{A differentiable rendering pipeline that integrates rasterization.}
    \label{fig:rasterization}
    \vspace{-0.4cm}
\end{figure}

Recent works~\cite{3dgs, adop, pulsar, nvdiffrast, pytorch3d} propose \emph{raster-based} differentiable rendering which enables high-speed rendering for 2D images (the forward pass) using rasterization techniques. Rasterization requires the scene to be composed of several discrete 3D geometric elements, or \emph{primitives} (e.g., triangles, points, ellipsoids). Each of these primitives is associated with shading information (e.g., color, opacity) and a position in space. Fig.~\ref{fig:rasterization} depicts how these primitives~\circled{1} (ellipsoids in this example) are rendered into 2D images~\circled{2}. Each pixel of the rendered image is thus influenced by a subset of the primitives in the scene. With differentiable rendering, all primitives are associated with a set of \emph{learnable parameters}~\circled{3} that are trained using gradient descent. For each training iteration (i.e., one image), the loss~\circled{4} is backpropagated~\circled{5} to compute the gradients for all the parameters associated with each primitive~\circled{6} (only the primitives that influence the current image). These parameters are updated with the computed gradients~\circled{7}, and the training iterations continue until convergence is achieved (i.e., the primitives are able to accurately represent the scene from all angles). 
A state-of-art work in raster-based differentiable rendering is 3D Gaussian Splatting~\cite{3dgs} which models the scene with 3D Gaussians (seen as ellipsoids) as the geometric primitives. 

\section{Motivation}
\label{sec:motivation_backward_runtime}

\begin{figure}[!htb]
\centering
    \vspace{-.3cm}
    \includegraphics[width=1\linewidth]{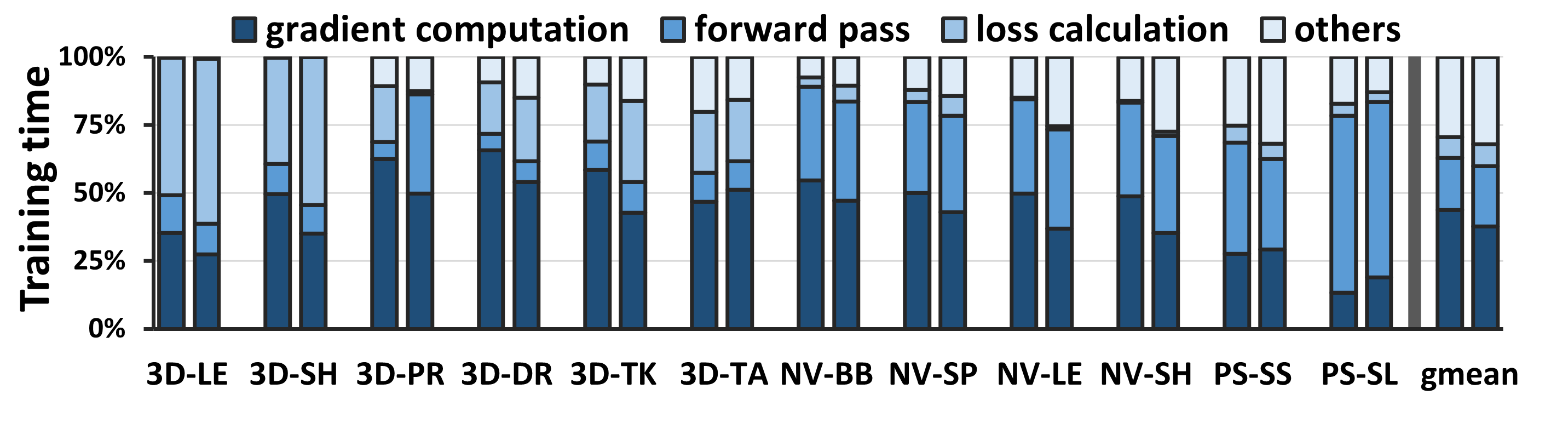}
    \vspace{-.6cm}
    \caption{Breakdown of training time on 4090 (left), 3060 (right).}
    \label{fig:backward_runtime}
    \vspace{-0.4cm}
\end{figure}

In this section, we profile important raster-based differentiable rendering workloads on the NVIDIA RTX 4090 GPU (methodology is described in~\cref{sec:methodology}).  Fig.~\ref{fig:backward_runtime} depicts the breakdown of training time, including the forward pass (during which an image is rendered from the model), loss calculation (which involves computing the difference between ground truth and rendered image), and the gradient computation (which involves computing and updating the loss gradient with respect to model parameters).
We make the following observations. First, we observe that on average 44\% (up to $66\%$) of the total execution time is spent on the gradient computation step and is thus a significant bottleneck in most workloads. Second, this bottleneck is most pronounced for workloads such as \texttt{3D-DR} and \texttt{3D-PL} (see \cref{sec:methodology}), taking up $65.8\%$ and $62\%$, of the overall runtime respectively. This is because \texttt{DR} and \texttt{PL} are real-world scenes that require a large number of primitives (i.e, a large model) for accuracy. 
The gradient computation time increases with scene size and complexity, whereas the forward pass and loss computation is independent of the scene complexity. Thus gradient computation becomes a bigger bottleneck  in more complex scenes.


\subsection{Atomic Reduction Bottleneck in the Gradient Computation}
\label{sec:motivation_shaderbwdimpl}

The input to the gradient computation kernel is a per-pixel list of primitives, where each list contains the IDs of primitives that influences the color of the corresponding pixel (discussed in~\cref{sec:bckg_diffrendering}). 
The gradient computation in the gradient computation step of differentiable rendering workloads is depicted in Fig.~\ref{fig:atomicimplementation}. 
Each thread (one per pixel) iterates through a list of its associated primitives (line 2, 3). Several intermediate conditions (like $cond1, cond2$ in lines $5$ and $9$) determine if the current thread contributes to each primitive's gradients. Each thread then computes the gradient contribution of the primitive's parameters ($grad_t x1, 
 grad_t x2, ...$). Finally, each thread performs an atomic add operation to atomically add its gradient contributions to the primitive's parameters (shown in lines 12-14). This operation needs to be atomic because multiple threads may update the same primitive's parameters.  

  

\begin{figure}[!htb]
\vspace{-0.4cm}
 \colorbox[RGB]{239,240,241}{
     \begin{minipage}{.9\linewidth}
\small
\begin{algorithmic}[1]

\Function{GradComputation}{prims\_per\_thread}

\State $tid \gets thread\_idx$
\Comment{Thread corr. to pixel}
\For{ $p : primitives[tid]$ }
\Comment{Iterate}
\If{ \Call{cond1}{} } 
    \State \textbf{continue;}
    \Comment{thread doesn't participate}
\EndIf
    \State ...    
\If{ \Call{cond2}{} }
    \State \textbf{continue;}
    \Comment{thread doesn't participate}
\EndIf
   
    \State ...    
\Comment{Gradient computation is done here}  
    \State \Call{AtomicAdd}{$p.grad\_x1$,  $grad_t x1$}
    \State \Call{AtomicAdd}{$p.grad\_x2$,  $grad_t x2$}
    \State \Call{AtomicAdd}{$p.grad\_x3$,  $grad_t x3$}
\EndFor

\EndFunction

\end{algorithmic}
\end{minipage}
}

\caption{Outline of the gradient computation step} 
\label{fig:atomicimplementation}
\vspace{-0.3cm}
\end{figure}

\begin{figure}[!htb]
\centering
    \centering
    \includegraphics[width=1\linewidth]{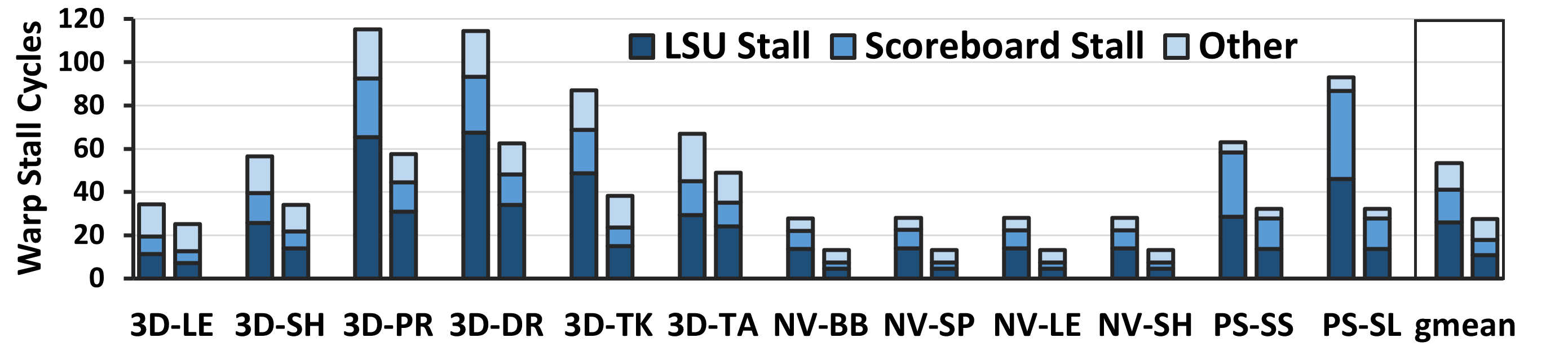}
    \caption{Breakdown of warp stalls on 4090(left), 3060(right).} 
    \label{fig:atomics_stalls}
    \vspace{-15pt}
\end{figure}

Given that each thread updates a number of primitives, each of which has many learned parameters, a massive number of atomic operations are generated (in the order of a few 10s to 100s of millions per iteration).
To evaluate the impact of this, we analyze the cycles during the gradient computation step when instructions are stalled from executing on two GPUs. Fig.~\ref{fig:atomics_stalls} depict the breakdown of the number of cycles a warp is stalled per instruction on the NVIDIA RTX 4090 and RTX 3060 GPUs using NVIDIA NSIGHT profiler~\cite{nsight_prof}. We make two observations. First, the LSU (load-store unit) stalls contribute to over 60\% of all stalls on average. The LSU stalls are caused due to the large number of memory requests (primarily atomic operations) to global memory from each sub-core (\cref{sec:bckg_atomicprocessing}). 
Second, the RTX 4090 GPU has more stalls in issuing instructions to the LSU compared to the RTX 3060.
This is because more recent GPUs have a higher SM to ROP unit ratio. In our experimental setup, the RTX 4090 has 5.14x more SMs than the RTX 3060 (144 SMs and 28 SMs respectively). However, the RTX 4090 only has about 3.6x more ROP units (176 ROP units versus 48 ROP units).

\vspace{-0.3cm}


\subsection{Key Observations}
\label{sec:motivation_observations}
We make the following observations from profiling atomic operations in the gradient computation step. 
\subsubsection{Observation 1} 
\begin{figure}[!htb]
    \vspace{-0.4cm}
    \begin{subfigure}{0.28\textwidth}
    \centering 
    \includegraphics[width=\linewidth]{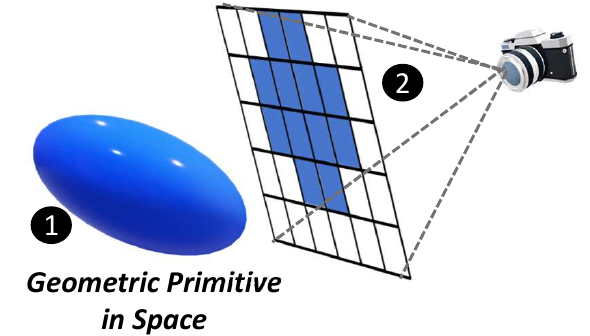}
    \vspace{-.5cm}
    \caption{Close-by pixels likely to be influenced by same primitive.}
    \label{fig:raster_locality}
    \end{subfigure}~
    \begin{subfigure}{0.21\textwidth}
    \centering
    \includegraphics[width=0.65\linewidth]{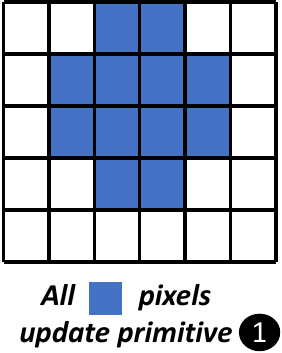}
    \vspace{-.2cm}
    \caption{Gradients of affected pixels are atomically aggregated}
    \label{fig:zoomed}        
    \end{subfigure}
    \caption{Close by threads (corresponding to pixels) update the parameters of the same primitive.}
    \label{fig:raster_locality_schematic}    
    \vspace{-0.3cm}
\end{figure}

\textbf{Threads within a warp are likely to update the same parameters.} 
Each primitive affects a region of pixels on the screen, called a ``fragment'' (\cref{sec:bckg_diffrendering}). As a result, close-by pixels that belong to the same fragment update the same primitive.  
Fig.~\ref{fig:raster_locality_schematic} shows how adjacent/close by pixels are part of the same fragment during rasterization. Fig.~\ref{fig:raster_locality} shows a  primitive in space \circled{1} rasterized onto a screen \circled{2} as seen from the camera indicated by the blue pixels during rendering. In the gradient computation step, each of these blue pixels affected will update the primitive's gradient. A zoomed in version of the captured image is shown in Fig.~\ref{fig:zoomed}. 

\begin{figure}[!htb]
    \centering
    \vspace{-0.4cm}
    \includegraphics[width=\linewidth]{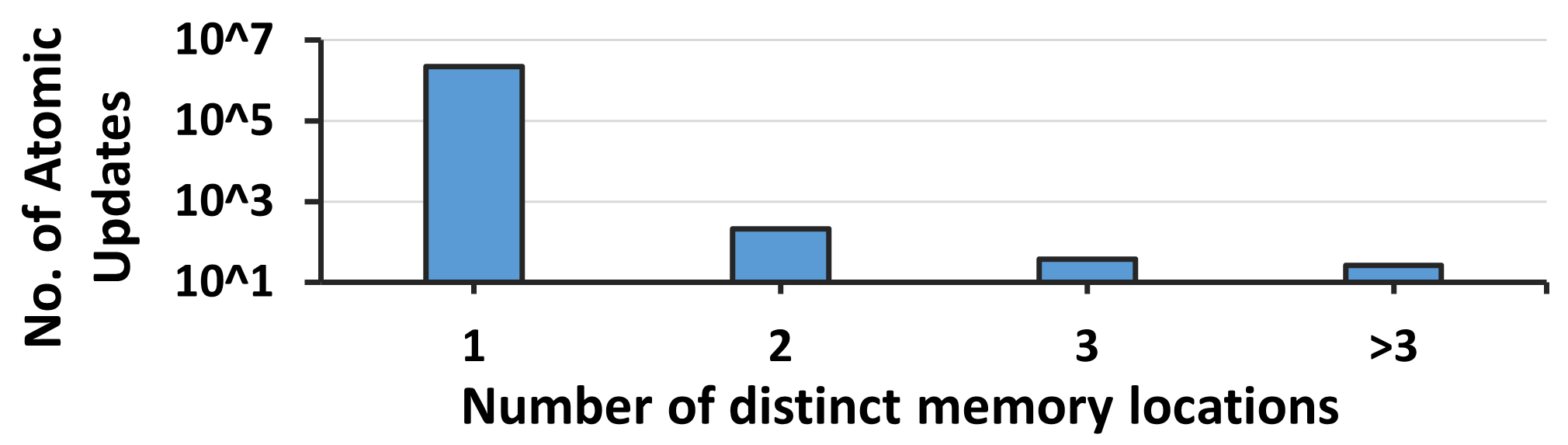}
    \vspace{-.6cm}
    \caption{Log-scale histogram of number of distinct memory locations updated by threads during gradient computation.}
    \vspace{-0.3cm}\label{fig:atomics_locality}
\end{figure}

Thus, threads within a warp (where each thread corresponds to one pixel and each warp corresponds to a local region of pixels) often compute the gradients for the parameters associated with the same primitive. 
These gradients are then atomically summed up across threads to update each parameter. 
We perform an experiment to determine the number of threads in each active warp that update the same parameters and thus, the same memory locations. Fig.~\ref{fig:atomics_locality} shows a histogram of the total number of memory locations that are atomically updated by each warp (at each loop iteration of Fig.~\ref{fig:atomicimplementation}). We observe that over 99\% of warps have all its threads update the same memory location.  

\subsubsection{Observation 2} 
\begin{figure}[!htb]
\begin{subfigure}{.242\textwidth}
    \centering
    \includegraphics[width=\linewidth]{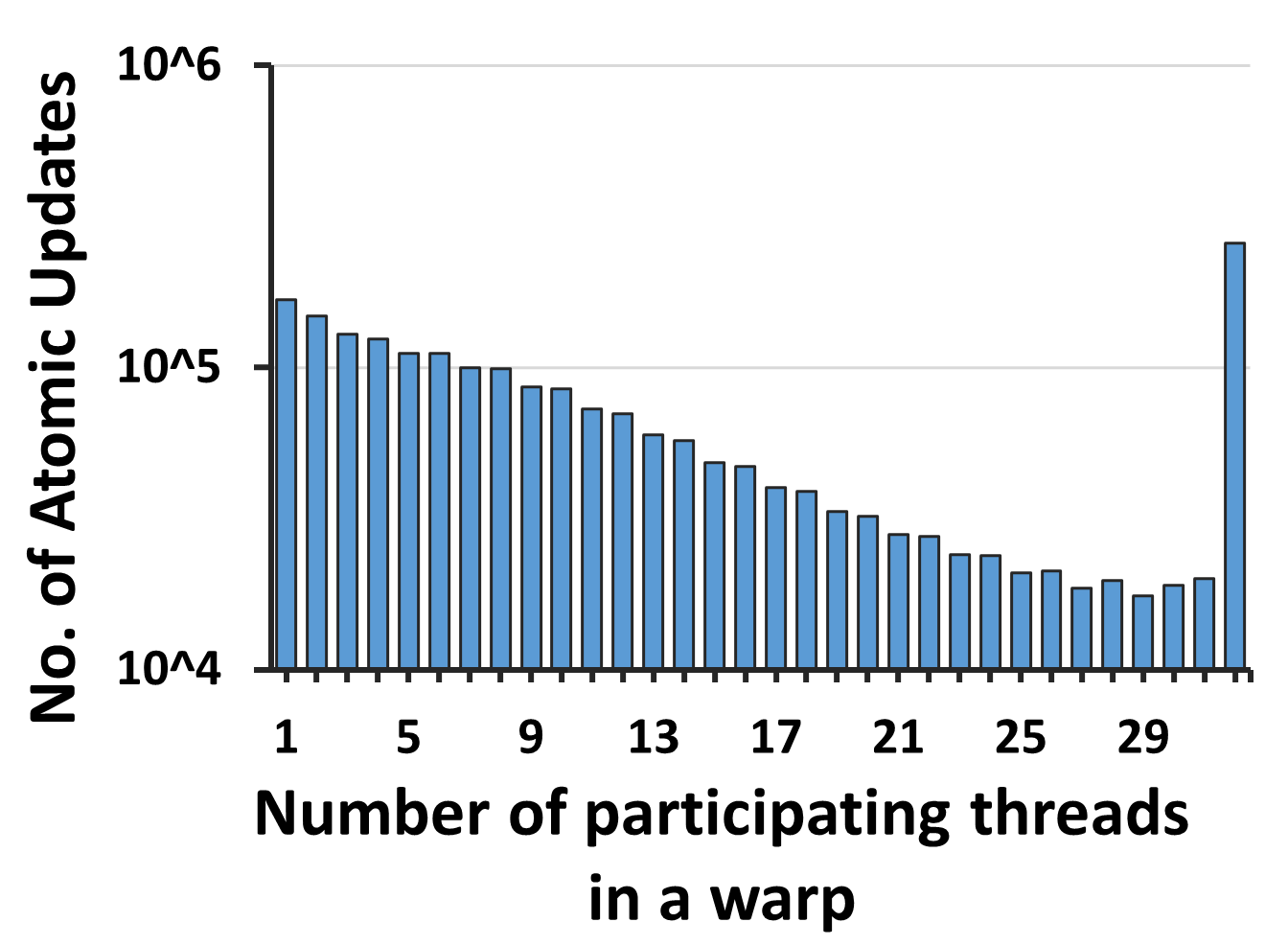}
    \caption{\texttt{3D-PR} workload}
\end{subfigure}~
\begin{subfigure}{.242\textwidth}
    \centering
    \includegraphics[width=\linewidth]{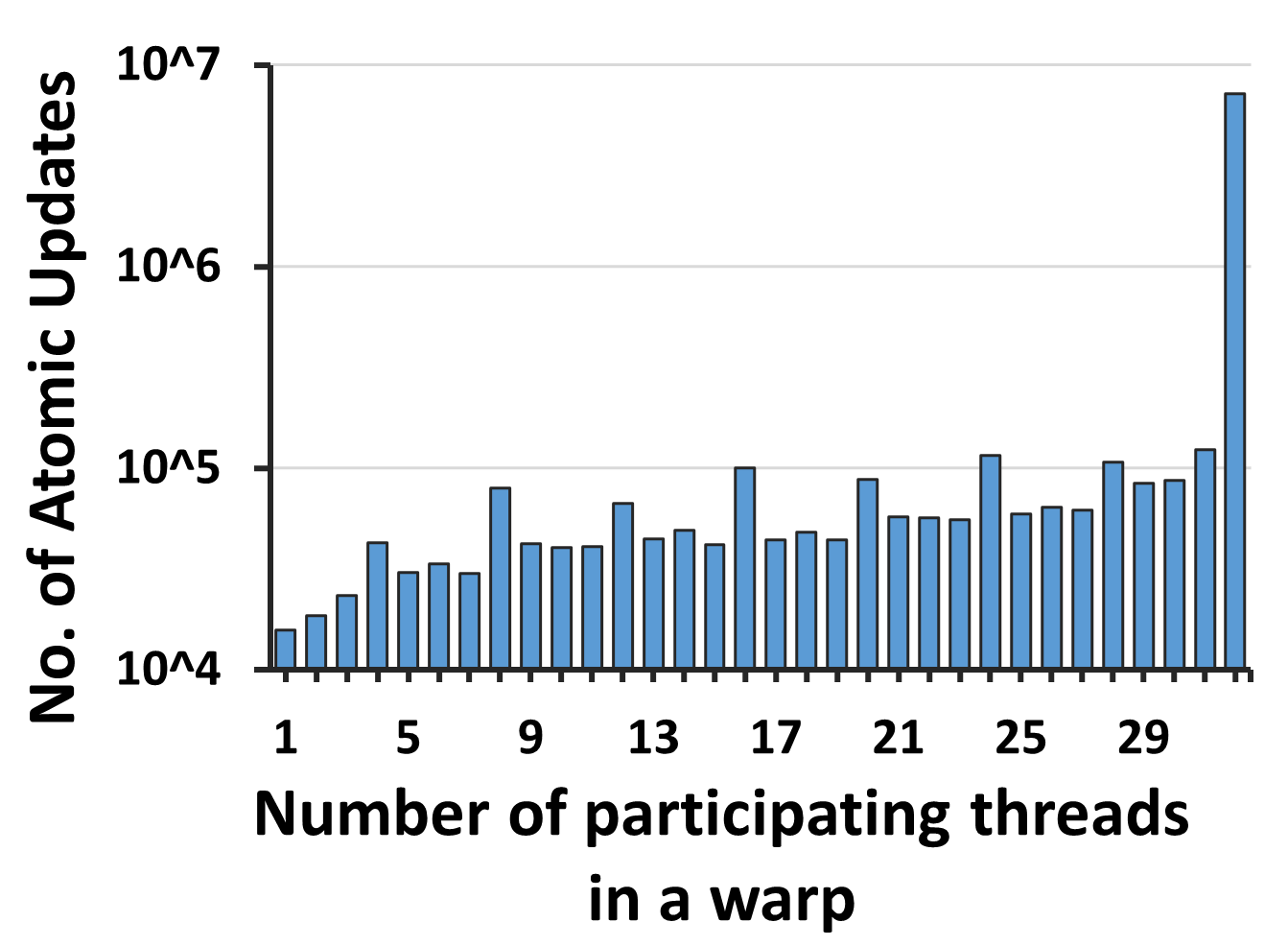}
    \caption{\texttt{NV-LG} workload}
\end{subfigure}
    \caption{Log-scale histograms of average number of active threads per warp participating in atomic updates.}
    \vspace{-0.3cm}
    \label{fig:atomics_unconvergent_warp_instructions}

\end{figure}

\textbf{Only a fraction of threads within a warp perform atomic updates at any given time.}
From Fig.~\ref{fig:atomicimplementation}, we see that the gradient computation step has certain dynamic conditions ($cond1, cond2, ...$) that cause some threads to skip the current iteration of gradient updates. Thus, only a fraction of all threads within a warp send out atomic requests in one iteration.
We measure the number of threads that typically participate in the atomic reduction in 
Fig.~\ref{fig:atomics_unconvergent_warp_instructions} for two different workloads \texttt{3D-PR} and \texttt{NV-LG} (refer to~\cref{sec:methodology} for workload-dataset configurations). We observe that there is significant variation in the number of threads in a warp that participate in one reduction. Thus, each warp contributes a different amount of traffic to the LSU and the ROP units. 

\vspace{-0.4cm}

In this work, our \textbf{goal} is to accelerate raster-based differentiable rendering applications by accelerating atomic operations that constitute a significant bottleneck in the gradient computation step. We describe in the next section how we leverage these observations to develop a streamlined and efficient technique to alleviate this bottleneck. 

\section{Approach}
We introduce \projname, a primitive that enables fast atomic reduction in applications that \textbf{(1)} generate a large number of atomic requests, thus overwhelming the hardware queues and compute units that process atomics, and \textbf{(2)} typically have most threads within an warp performing atomic updates to the same memory locations. 

The \textbf{key ideas} behind \projname is to (i) leverage the intra-warp locality in atomic updates to perform warp-level reduction in the SM itself using registers, and (ii) distribute atomic computation between the SM and L2 ROP units to enable high throughput atomic reduction. 
We propose a SW only implementations of \projname that leverages existing warp-level primitives to implement reduction at warp level.



\subsection{Design Challenges of \projname}
\label{sec:design_challenges}


\textbf{Challenge \circled{1}: All threads in warp may not generate atomic updates.}  Only a subset of threads in a warp typically generate atomic updates at any given time (as discussed in~\cref{sec:motivation_observations}). Existing warp-level primitives thus cannot be directly used to perform warp-level reduction for differentiable rendering workloads. This irregularity poses challenges in developing an efficient implementation of warp-level reduction at the core for both hardware and software approaches. 

\textbf{Challenge \circled{2}: Dynamic scheduling of atomic computation between the core and L2.} To meet the high throughput requirements for atomic computations in differentiable rendering, it is critical to effectively use both existing ROP units at the L2 as well as the proposed warp-level reduction at the core. Thus, \projname must automatically perform this scheduling efficiently at runtime based on the utilization of the atomic units at the core and L2. 

\subsection{Key Components of \SWprojname}
\label{sec:approach_sw}

\SWprojname is implemented and exposed to the programmer as a function call that can be inserted in GPU code.  
We now describe how we implement \SWprojname using existing instructions and warp-level primitives. 

\begin{figure}[!htb]
    \centering
    \includegraphics[width=.65\linewidth]{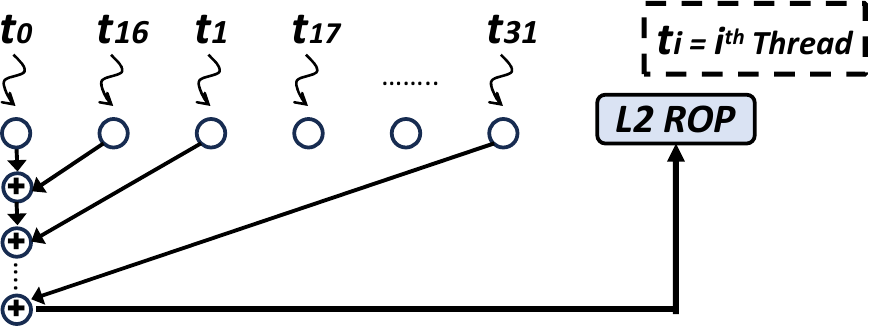}
    \caption{Serialized reduction implementation overview}
    \vspace{-.3cm}
    \label{fig:serialized_overview}
\end{figure}

\textbf{Warp-level Reduction (Challenge \circled{1}).}  
We propose two approaches to perform warp-level reduction that addresses Challenge \circled{1}, each of which has different tradeoffs. These approaches are outlined below:

\textbf{(1) Serialized Reduction:} Within each warp, we first determine a set of threads that atomically update the same parameter (and thus, memory location). One thread out of this group then iterates through all the gradients (one from each thread) and this is depicted in Fig.~\ref{fig:serialized_overview}. The accumulated result is then added to the parameter using an regular atomic add operation. The serial nature of this approach is inefficient. However, when the warp has threads updating multiple parameters, the reductions can be parallelized. We develop an efficient implementation of serialized reduction by batching updates to all parameters associated with the primitive, discussed in~\cref{sec:dd_serialized}. 
\vspace{-0.4cm}

\begin{figure}[!htb]
    \centering
    \vspace{-.3cm}
    \includegraphics[width=.65\linewidth]{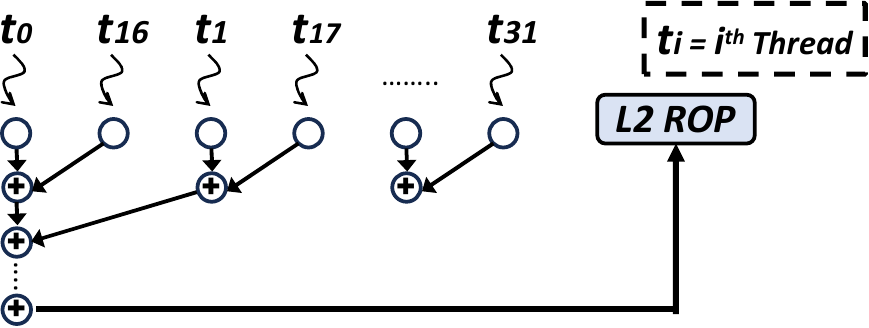}
    
    \caption{Reduction-tree/butterfly reduction overview}
    \vspace{-.3cm}
    \label{fig:bfly_overview}
\end{figure}

\textbf{(2) Butterfly Reduction:} 
    Fig.~\ref{fig:bfly_overview} shows how butterfly reduction is performed for threads in a warp. We first check whether all the threads in a warp update the same primitive. If so, we use a reduction tree to sum the gradients.
    For this implementation to work, it requires all threads to be active, or for threads that are inactive, we must add a $0$ value. This introduces some redundant computation. Thus, the programmer has to ensure there is no control flow divergence and  all threads are active, and assign 0-value atomic update to threads that originally did not participate in the gradient summation. Butterfly reduction is most efficient when there is only one parameter being updated by the warp and most threads are active (less redundant updates).

\textbf{Scheduling Atomic Updates Between Core and L2 ROP (Challenge \circled{2}).} 
As discussed in~\cref{sec:motivation_observations}, the amount of contention at the LSU that is contributed to by each warp depends on the number of active threads producing atomic requests. Additionally, the active thread count is also a measures the amount of reduction ``work'' to be done in the SM (if the atomic update is scheduled for warp-level reduction).
To address Challenge \circled{2}, we determine whether the atomic updates should be performed using a warp-level reduction at the core or at the L2 ROPs, by comparing the number of threads in the warp that actively update one parameter against a \emph{predefined threshold}. 
We call this threshold the balancing threshold, as it balances the atomic computation between ROP units and the SMs. This scheduling is performed for each set of threads in a warp that updates one parameter. 
The optimal \emph{balancing threshold} depends on the amount of contention in the atomic units. This in turn depends on the following factors:
\begin{itemize}
    \item \textbf{Dataset (scene) and workload:} The number of atomic updates depends on factors such as the camera resolution, model architecture, and the size/complexity of the scene being learned.
    \item \textbf{GPU architecture:} The ratio of SMs to ROP units impacts the contention at cores and ROP units. 
    \item \textbf{Reduction method used:} The choice of using the butterfly or serial reduction methods also affects the contention at the atomic units.
\end{itemize}
Due to the complexity in determining the threshold analytically, we treat the balancing threshold as a hyperparameter that needs to be tuned for each workload. We discuss in detail how we used the balancing threshold in~\cref{sec:design_main} and evaluate the impact of this hyperparameter in~\cref{sec:sw_speedup}.



\section{Detailed Design}
\label{sec:design_main}


\subsection{Design of \SWprojname}
\label{sec:sw_reduction}


\subsubsection{\textbf{\SWprojname with Serialized Reduction (SW-S)}}
\label{sec:dd_serialized}  As discussed in~\cref{sec:approach_sw}, this implementation performs the warp-level reduction serially. It is exposed to the programmer as a function call that is invoked during gradient computation, directly replacing the atomic instructions in Fig.~\ref{fig:atomicimplementation} (lines 14-16) and is called by all threads. The function's implementation is provided in Fig.~\ref{alg:redserial}. It takes as input: the primitive to be updated by the thread, the primitive's parameters, and the gradients generated by the calling thread for all the primitive's parameters. Each thread determines how many other threads in the warp are updating the same primitive (done using \texttt{\_\_match\_any\_sync}, line 10). If this is less than the balancing threshold, the function simply sends the original atomic updates (lines 36-38), and thus uses the ROP units for reduction.
Otherwise, for each primitive, a leader thread is identified (the thread in the warp with the lowest lane ID, line 18). This thread serially accumulates gradients across all active threads in a warp for all the parameters associated with the primitive (line 22-30). The leader thread thus skips inactive threads and threads that update other primitives. It then generates one atomic update instruction per parameter, that is sent in a normal manner to the ROP unit (line 31-34).

\textbf{Limitations:} The primary limitation is the inefficient serial reduction with execution time proportional to the number of active threads per primitive. This also involves additional control flow overheads (lines 16,24,26,27,32,33,37). 

\begin{figure}[!htb]
\vspace{-0.4cm}
\begin{lstlisting}
// Input - primitive index idx, pointers to 
parameter gradients, values to be accumulated, 
balancing threshold
template<typename ATOM_T>
void reduce_serial(int idx, ATOM_T** ptr, 
  ATOM_T *val, int num_params, int balance_thr) {
  /* a mask of threads in current warp updating
  the same primitive and a count of how many 
  threads in this mask.*/
  int same_mask = "match_any"(idx);
  int same_ct = "popc"(same_mask);

  /* if number of threads updating current 
  primitive exceeds balance threshold, perform
  serialized warp level reduction */
  if (same_ct >= balance_thr) {
    // thread with lowest id becomes the leader
    int leader = "ffs"(same_mask) - 1;
    /* leader does not fetch from itself */
    same_mask &= ~(1 << leader);
    
    /* leader fetch and accumulate all parameters
    from threads updating the same primitive */
    while (same_mask) {
      int src_lane = __ffs(same_mask) - 1;
      if (laneId==leader || laneId==src_lane)
        for (int i = 0; i < len; ++i) 
          val[i] += __shfl(val[i], src_lane);
      same_mask &= ~(1 << src_lane);
    }
    /* leader sends an atomicAdd per parameter */
    if (laneId == leader)
      for (int i = 0; i < num_params; ++i)
        atomicAdd(ptr[i], val[i]);
  } else {
    /* balance threshold not met, update normally */
    for (int i = 0; i < num_params; ++i) 
      atomicAdd(ptr[i], val[i]);
  }
}
\end{lstlisting}
\vspace{-0.3cm}
\caption{\textbf{CUDA implementation of SW-S routine.}}
\vspace{-16pt}
\label{alg:redserial}

\end{figure}

\subsubsection{\textbf{\SWprojname with Butterfly Reduction (SW-B)}}
\label{sec:dd_butterfly}
As discussed in~\cref{sec:motivation_observations}, over 99\% of warps in many workloads have all active threads update the same primitive's parameters. In these cases, a parallelized reduction tree can be used for fast warp-level reduction. We propose an efficient implementation that requires that (1) all threads in a warp update the same primitive and (2) all threads actively participate in the reduction. The programmer can use SW-B only if the first condition is met. To ensure all threads participate in the reduction, the previously inactive threads must now be made to generate zero value gradient updates. 
Fig.~\ref{alg:butterfly} presents our implementation. This function is similar to SW-S but also receives an input variable that indicates if the thread was active and is updating a non-zero value gradient. This variable is used to determine if the number of active threads in the warp is greater than the balancing threshold (using \texttt{\_\_ballot\_sync}, line 14). If so, a butterfly reduction is performed using \texttt{shfl} instructions (line 20-22). 

\textbf{Limitations:} SW-B adds redundant computation by making inactive threads perform zero value gradient updates, making reduction for warps with many inactive threads inefficient. Using SW-B also requires changes to the kernel code demonstrated with an example in Fig.~\ref{fig:bflyimplementation}, where the code is transformed to ensure all threads participate in the reduction. This transformation can be non-trivial in some applications. 


\begin{figure}[!htb]
\vspace{-0.3cm}
\begin{lstlisting}
// Input - primitive index idx, pointers to 
parameter gradients, values to be accumulated, 
balancing threshold, a boolean that indicates
whether current thread is participating
template<typename ATOM_T>
void reduce_bfly(int idx, ATOM_T** ptr,
  ATOM_T *val, int num_params, int balance_thr, 
  bool was_active) {
  /* reduction only performed when all threads
  are updating the same primitive */
  bool all_same = "match"(idx) == 0xffffffff;
  
  // number of threads making nonzero updates
  int same_ct = "ballot"(was_active); 

  /* number of threads updating current 
  primitive exceeds balance threshold, perform
  warp level butterfly reduction */
  if (all_same && same_ct >= balance_thrsh) {
    // parallel butterfly reduction tree
    for(int offs = 16; offs >= 1; offs /= 2)
        val[i] += "shfl_down"(val[i], offset);
    // first thread has accumulated gradients
    // send an atomicAdd per parameter
    if (laneId == 0) 
      atomicAdd(ptr[i], val[i]);
  } else if (was_active) {
    /* if balance threshold is not met or 
    butterfly reduction is ineligible, update 
    gradients normally with atomic operations */
    for (int i = 0; i < num_params; ++i)
      atomicAdd(ptr[i], val[i]);
  }
}
\end{lstlisting}
\vspace{-0.3cm}
\caption{\textbf{CUDA implementation SW-B routine.}}
\vspace{-.2cm}
\label{alg:butterfly}
\end{figure}

\begin{figure}[!htb]
\footnotesize
 \colorbox[RGB]{239,240,241}{
     \begin{minipage}{0.95\linewidth}
\begin{algorithmic}[1]

\Function{GradComputeBFLY}{prims\_per\_thread}

\State {tid = thread\_idx}
\State {prims\_per\_thread = primitives[tid]}

\For{ p in prims\_per\_thread }

\State {{\color{red}\textbf{was\_active = true;}} \textit{\quad // active by default}}
\If{ \Call{cond1}{} }
    \State {\textit{// instead of skipping, mark inactive status}}
    \State {\color{red}\textbf{was\_active = false;}}
\EndIf
    \State ...    
\If{ \Call{cond2}{} }
    \State {\textit{// instead of skipping, mark inactive status}}
    \State {\color{red}\textbf{was\_active = false;}}
\EndIf
    \State ...    
    \If{ not \textbf{was\_active} }
        \State {\textit{// thread was inactive, assign zero gradients}}
        \State {{\color{red}\textbf{grad$_{x1, ...xN}$ = 0}}}
    \EndIf
    \State g\_ptrs = array[p.grad$_{x1, ...xN}$]
    \State g\_vals = array[grad$_{x1, ...xN}$]
    \State {\textit{// pass inactive status to SW-B routine}}
    \State \Call{red\_bfly}{p, g\_ptrs,  g\_vals, N, {\color{red}\textbf{was\_active}}}
    \State ...    

\EndFor

\EndFunction

\end{algorithmic}
\end{minipage}
}

\caption{\textbf{Outline of a modified gradient computation kernel (Fig.~\ref{fig:atomicimplementation}) that integrates the SW-B primitive.}} 
\vspace{-0.5cm}
\label{fig:bflyimplementation}


\end{figure}




\subsubsection{\textbf{Determining Balancing Threshold}}
\label{sec:manualtuning}
The balancing threshold significant impacts speedups (evaluated in~\cref{sec:eval_balance}) and needs to be tuned for best results.  
The balancing threshold has only 32 possible values ($0-31$), and the gradient compute kernel is called 100000s of times during training. Thus, we present a simple method to automatically tune the threshold: We execute one iteration of the gradient computation kernel using all 32 values of the threshold and select the value that provides the largest speedup. We repeat this profiling every $N$ iterations (2000 in our evaluation). This profiling step adds a negligible amount of overhead, as the profiling iterations are significantly fewer than the training iterations.





\section{Methodology}

\label{sec:methodology}

\textbf{Evaluation Platform.} 
We implement and evaluate \projname-SW on real hardware setups with an Intel Core i9 13900KF CPU 
and the NVIDIA RTX4090 and RTX3060 GPUs. 


\textbf{Workloads.} We evaluate \projname using widely used raster-based differentiable rendering applications, described below:
\begin{itemize}[leftmargin=10pt]
    \item \textbf{\texttt{3DGS}}: 3D Gaussian Splatting~\cite{3dgs} represents the scene with a set of 3D Gaussians. Each Gaussian is associated with view dependent radiance and is learned during the differentiable rendering training process.
    
    \item \textbf{\texttt{NvDiffRec}}: Nvdiffrec~\cite{nvdiffrec} is a large project used for various differentiable rendering tasks. In our evaluation, we use differentiable rendering to learn the parameters of specular cubemap texture from a set of mesh images. 
    
    \item \textbf{\texttt{Pulsar}}: Pulsar~\cite{pulsar} is a recent work for 3D scene reconstruction, that represents the scene with a set of spheres with an efficient sphere rasterizer. This implementation is incorporated into Pytorch3D~\cite{pytorch3d}, a widely used framework for differentiable rendering. 


\end{itemize}

We evaluate our approach using the datasets listed in Table~\ref{table:datasets}. For \texttt{pulsar}, we use two synthesized datasets comprising 3D spheres (\texttt{PS-SS} and \texttt{PS-SL}).

\begin{table}[!htb]
 \centering
  \caption{Workloads and datasets}
  \label{table:datasets}
  \begin{tabular}{|l|c|l|}
    \hline
     \multirow{2}{*}{\textbf{Workloads}}     & \textbf{Dataset}      & \multirow{2}{*}{\textbf{Dataset name}} \\
                                             & \textbf{identifier}   &  \\
    \hline
    \multirow{5}{*}{\texttt{3DGS} (\texttt{3D})} &  \texttt{LE} & NerfSynthetic-Lego~\cite{nerf}       \\
                                   &  \texttt{SH} & NerfSynthetic-Ship~\cite{nerf}       \\
                                   &  \texttt{PR}   & DB COLMAP Playroom~\cite{playroom}       \\
                                   &  \texttt{DR}   & DB COLMAP DR. Johnson~\cite{drjohnson}       \\
                                   &  \texttt{TK} & Tanks and Temples-Truck~\cite{tandt}   \\
                                   &  \texttt{TA} & Tanks and Temples-Train~\cite{tandt}   \\
                                   
    \hline
    \multirow{2}{*}{\texttt{NvDiffRec} (\texttt{NV})} &  \texttt{BB} &  Keenan Crane 3D model - Bob~\cite{crane}   \\
                                    &  \texttt{SP} &  Keenan Crane 3D model - Spot~\cite{crane}   \\
                                   &  \texttt{LE} & NerfSynthetic-Lego~\cite{nerf}       \\
                                   &  \texttt{SH} & NerfSynthetic-Ship~\cite{nerf}       \\
    \hline
    \multirow{2}{*}{\texttt{pulsar} (\texttt{PS})} &  \texttt{SS}   & Synthetic Spheres - Small  \\
                                     &  \texttt{SL}   & Synthetic Spheres - Large       \\

    \hline


  \end{tabular}
  \vspace{-0.4cm}
\end{table}

\section{Evaluation}

We evaluate 3 different \projname configurations: (i) \texttt{SW-B-X}: an implementation of \SWprojname using butterfly reduction, with balancing threshold $X$. (ii) \texttt{SW-S-X}: an implementation of \SWprojname using serialized reduction, with balancing threshold $X$. 
We refer to the configurations of \texttt{SW-B-X} and \texttt{SW-S-X} with the best performing balancing threshold as \texttt{SW-B} and \texttt{SW-S} respectively.
We also compare our work against: (iv) \texttt{CCCL} uses the existing NVIDIA CCCL library~\cite{cccl, cub}  to perform warp-level reductions. We test DISTWAR on real hardware: (i) \texttt{4090}: NVIDIA RTX 4090 GPU and (ii) \texttt{3060}: NVIDIA RTX 3060 GPU.

\subsection{Performance analysis}
\label{sec:sw_speedup}
\begin{figure}[!htb]
    \centering
    \vspace{-0.3cm}
\begin{subfigure}{0.49\textwidth}
    \includegraphics[width=\linewidth]{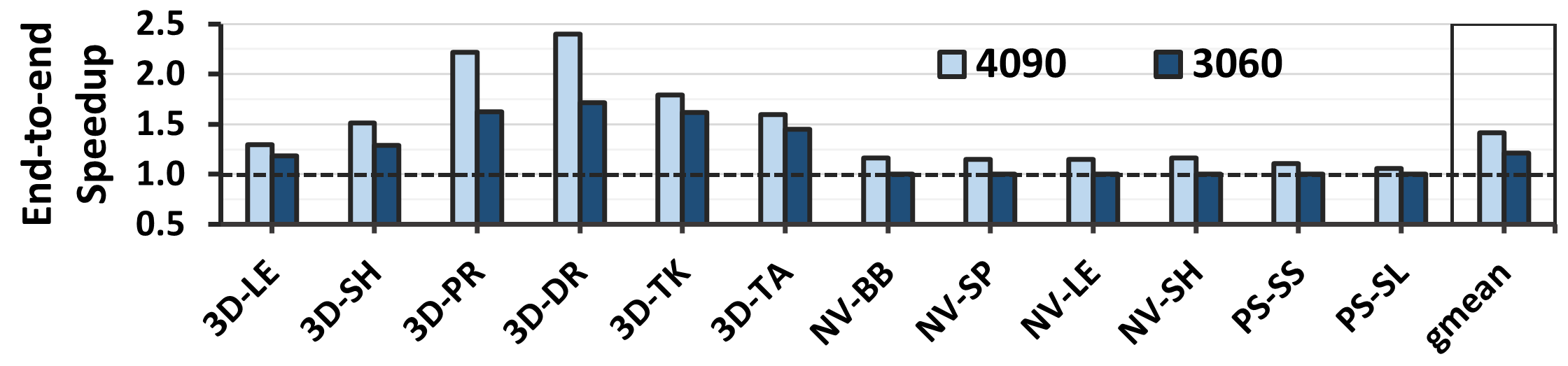}
\end{subfigure}
\begin{subfigure}{0.49\textwidth}
    \centering
    \includegraphics[width=\linewidth]{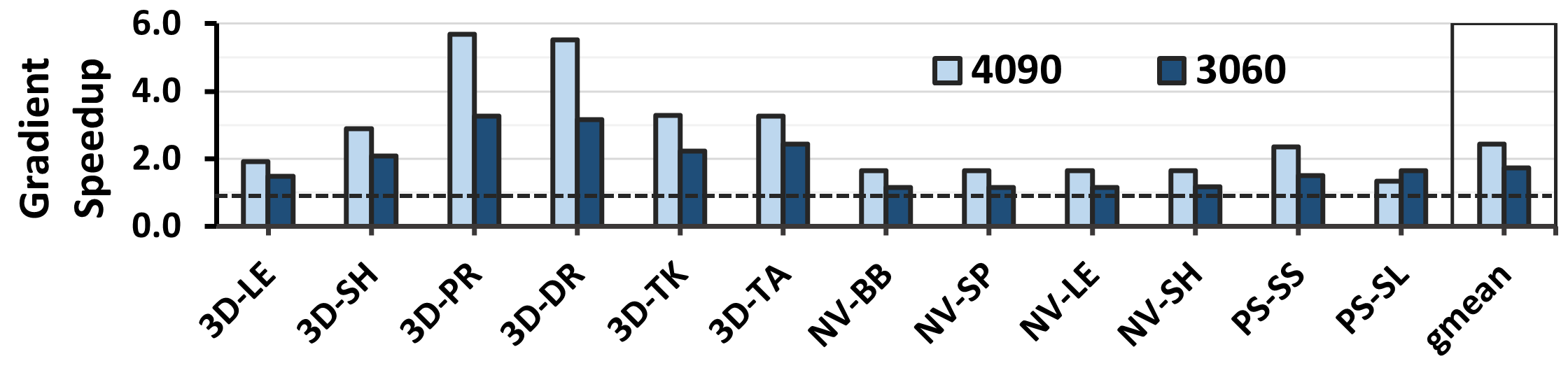}
    \vspace{-.65cm}
\end{subfigure}

\caption{End-to-end and gradient computation speedup normalized to baseline on \texttt{4090} and \texttt{3060}.}
\vspace{-0.2cm}
\label{fig:sw_speedup}
\end{figure}

Fig.~\ref{fig:sw_speedup} shows the normalized speedup for end-to-end runtime (including the forward pass) and the normalized speedup for the gradient computation alone. Speedups depicted in both graphs are normalized to baseline. 
Fig.~\ref{fig:warp_stall_sw} shows the average number of warp stalls per instruction and its breakdown on \texttt{4090} and \texttt{3060}.
We make the following observations:

First, both \texttt{SW-B} and \texttt{SW-S} are able to significantly outperform the baseline on average on both GPUs.
For the gradient computation, \texttt{\SWprojname} achieves an average speedup of $2.44\times$ (up to $5.7\times$) on \texttt{4090}, and $1.74\times$ (up to $3.27\times$) on \texttt{3060}.
For the entire differentiable rendering pipeline, \texttt{\SWprojname} achieves an average speedup of $1.41\times$ on \texttt{4090} (up to $2.4\times$) , and $1.21\times$ (up to $1.71\times$) on \texttt{3060}.


Second, we observe higher speedups on \texttt{4090}, compared to \texttt{3060}. This is because the atomic processing bottleneck is more pronounced on \texttt{4090} that has a lower ROP to SM ratio (containing 144 SMs and 176 ROP units versus 28 SMs and 48 ROPs in the 3060). 
Third, in our evaluation, \texttt{SW-B} performs as well as or much better than \texttt{SW-S}, which performs the reduction serially. However, there are some workloads (\texttt{PS-SS} and \texttt{PS-SL}) that cannot use \texttt{SW-B} because it was difficult to eliminate thread divergence which is a requirement for butterfly reduction (\cref{sec:dd_butterfly}).  
Fourth, we observe significantly higher speedups on \texttt{3D-PR} and \texttt{3D-DR}. This is because the datasets \texttt{PR}, \texttt{DR} are large-scale, photorealistic scenes that require many more geometric primitives (gaussians for \texttt{3D}) for accurate scene representation compared to the smaller scenes. This leads to a larger number of parameters that need to be atomically updated during gradient computation, making the atomic bottleneck more pronounced. 
Finally, we observe smaller end-to-end speedups in \texttt{NV} and \texttt{PS}. \texttt{NV} has much fewer warp stalls compare to \texttt{3D} in the baseline application (Fig.~\ref{fig:atomics_stalls}). This leads to a less contended LSU, which diminishes the speedups achieved by \texttt{\SWprojname}. In \texttt{PS}, even though the LSU is heavily contended during gradient computation (Fig.~\ref{fig:atomics_stalls}), the gradient computation is not the main bottleneck (Fig.~\ref{fig:backward_runtime}).

\subsection{Impact of the Balancing Threshold}
\label{sec:eval_balance}
\begin{figure}[!htb]
    \centering
    \vspace{-0.25cm}
    \includegraphics[width=1.0\linewidth]{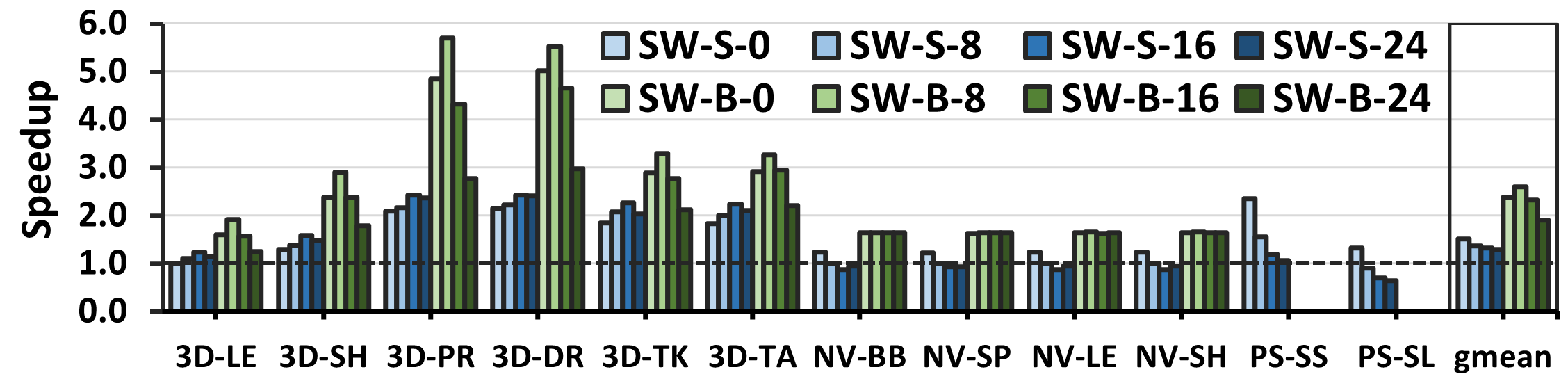}
    \caption{Sensitivity of \texttt{\projname-SW-S} and \texttt{\projname-SW-B} to the balancing threshold $X$. \texttt{SW-B} is cannot be used for \texttt{PS-SS} and \texttt{PS-SL}.}
    \label{fig:sw_balance_4090}
    \vspace{-0.1cm}
\end{figure}

In Fig.~\ref{fig:sw_balance_4090}, we depict the sensitivity of \texttt{\projname-SW-S} and \texttt{\projname-SW-B} speedups to the balancing threshold $X$ for the gradient computation on \texttt{4090}.
We make two observations. First, the best performing balancing threshold varies across workloads and datasets. For most workload configurations, we achieve the highest speedup when the balancing threshold parameter is set to ensure that the atomic updates are distributed between the ROP units and the SMs for both \texttt{SW-S} and \texttt{SW-B}. Thus setting $0$ or $24$ as the balancing threshold leads to contention in either the sub-core reduction unit or the ROP units respectively in these workloads. Second, in some workloads (\texttt{NV-BB, NV-SP, NV-LE, NV-SH, PS-SS, PS-SL}), choosing sub-optimal balancing thresholds can even lead to slowdowns. This is because in some compute-bound workloads, the additional instructions required to perform warp-level reduction can incur significant overheads. In these cases, balancing thresholds that favor the ROP unit should be chosen.



\subsection{Reduction in Stalls}
\begin{figure}[!htb]
\vspace{-0.3cm}
\centering
\includegraphics[width=1\linewidth]{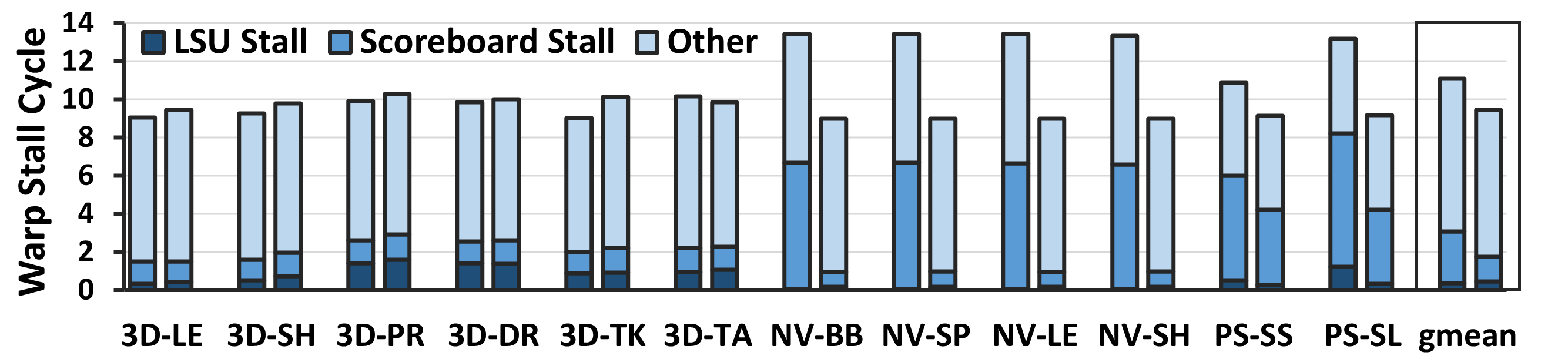}
\caption{Breakdown of warp stalls during gradient computation using \texttt{\SWprojname} on \texttt{4090} (left) and \texttt{3060} (right).}
\label{fig:warp_stall_sw}
\vspace{-0.3cm}

\end{figure}

To analyze where the performance speedups come from, we measure the number of stall cycles per instruction in Fig.~\ref{fig:warp_stall_sw} using the NVIDIA Nsight Compute~\cite{ncu} profiling tool. We observe significantly fewer overall stalls per instruction across all workloads compared to baseline (Fig.~\ref{fig:atomics_stalls}): 10.25 cycles versus 38.26 cycles on average. This is a result of significantly fewer stalls due to atomics (LSU stalls). 








\subsection{Comparing Against CCCL Library Implementation}
\label{sec:results_cccl} 
\begin{figure}[!htb]
    \centering
    \includegraphics[width=1\linewidth]{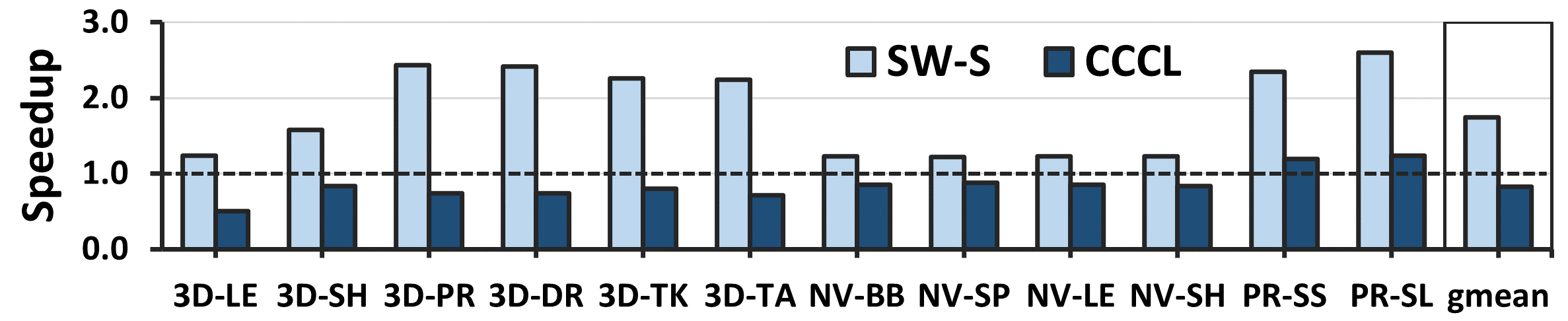}
    \caption{Gradient computation speedup of \texttt{\projname-SW-S} over \texttt{CCCL} on \texttt{4090}, normalized to baseline.}
    \vspace{-.5cm}
    \label{fig:cccl_comparison}
\end{figure}

\begin{figure}[!htb]
    \centering
    \includegraphics[width=1\linewidth]{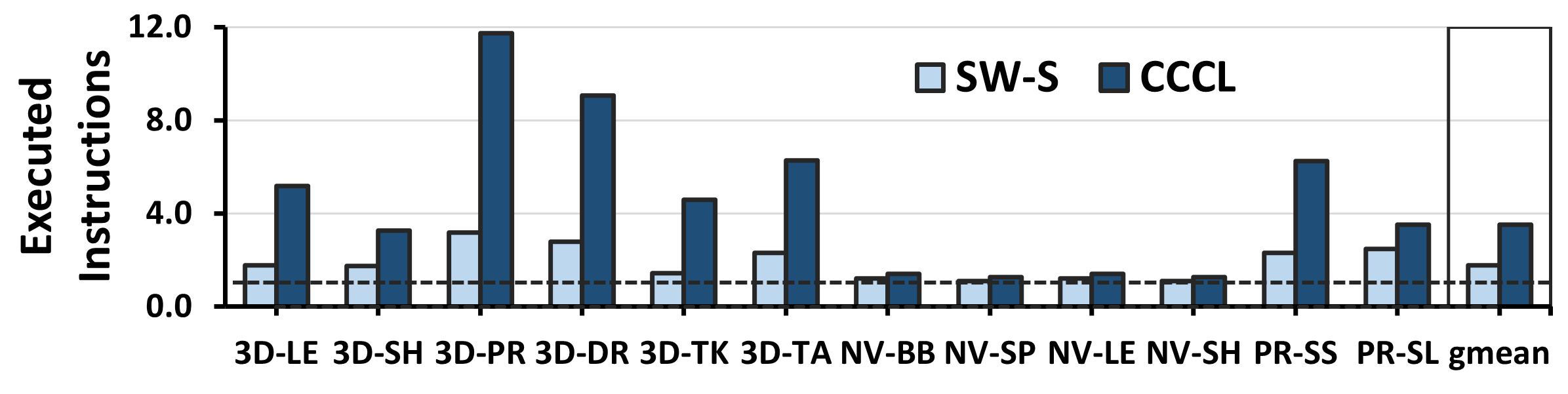}
    \caption{Normalized number of executed warp instructions of 
    \texttt{\projname-SW-S} over \texttt{CCCL} on \texttt{4090}.}
    \vspace{-.4cm}
    \label{fig:cccl_exec_insn}
\end{figure}

In Fig.~\ref{fig:cccl_comparison}, we compare against the state-of-art approach for software warp-level reduction, the NVIDIA CCCL Library~\cite{cccl}. We depict the speedup normalized over baseline and the gradient computation for \texttt{SW-S} and \texttt{CCCL} respectively on \texttt{4090}. 
We observe that using \texttt{CCCL} for warp-level reduction leads to an average slowdown of about $20\%$ across all workloads. \texttt{CCCL} is inefficient for differentiable rendering workloads because \emph{(i)} it performs a reduction operation for each parameter, while \SWprojname batches all parameters in a primitive (\cref{sec:dd_serialized}); and \emph{(ii)} does not perform distribution of atomic computation between the SMs and ROP units. \texttt{CCCL} also cannot be directly used when all threads in a warp are not active, requiring further addition of instructions. 
Fig.~\ref{fig:cccl_exec_insn} shows the significantly larger numbers of  instructions executed by \texttt{CCCL} compared to \SWprojname due to these inefficiencies. 

\vspace*{-0.2cm}



\section{Related Work}

\label{sec:related_work}

To our knowledge, this is the first work to \emph{(i)} characterize emerging raster-based differentiable rendering workloads and identify the atomic operations to be a key performance bottleneck; and \emph{(ii)} propose an efficient method to leverage warp-level reduction and existing atomic units to accelerate the processing of atomic updates in GPUs. 

\textbf{Accelerating differentiable rendering.}
Recent works have proposed software techniques~\cite{dvgo, instantngp, mobilenerf, plenoxels,nerfacc} as well as hardware accelerators~\cite{neurex, instant3disca, gennerf, artist} to accelerate both training and rendering for neural radiance fields (NeRF)~\cite{nerf,instantngp} methods. These works target one class of differentiable rendering applications typically used for scene reconstruction. With NeRFs, the primary bottleneck is due to the large number of computations and memory accesses required to both train and render a model with a large number of learned parameters. Raster-based differentiable rendering methods significantly reduce the number of computations required, making it a powerful and popular approach. However, it is still bottlenecked by atomic operations during training which we tackle in this work. NeRF methods also have atomic contention during training that is not addressed by prior work, but atomics only constitute a secondary bottleneck in these workloads. To our knowledge, this is the first work to characterize and propose techniques to accelerate raster-based differentiable rendering workloads. 

\textbf{Accelerating atomics in GPUs using SM-level buffering.}
Remote memory operations (RMOs)~\cite{rmo,t3ermo,networkrmo} process atomic operations by adding hardware to do computations near shared data caches. Modern GPUs use an RMO-architecture to process atomic operations~\cite{fermiarch}, as they offer a convenient way to process atomics without cache coherence protocols.
However, this can lead to additional memory traffic and prior work~\cite{lab} proposes to perform some atomic updates at the SM to reduce contention at the ROP units by buffering updates at the L1. However, this approach is not effective when the workload produces a massive number of atomic updates that overwhelm the LSU before the updates can be buffered. In comparison, \projname leverages the intra-warp locality in atomic updates seen in differentiable rendering workloads to perform warp-level reduction using registers at the SM. This approach significantly reduces the number of atomic updates sent to the LSU and the partitioning approach dynamically leverages both the ROP units and the SMs to enable high throughput processing of atomics.

Deterministic atomic buffering~\cite{dab} is another approach that buffers atomic requests in the SM to maintain the determinism in the order of atomic execution, but does not aim to improve speed of atomic updates. 
Using a modified memory consistency model for GPUs that allows threads to synchronize at the L1 enables buffering of atomic operations at the SM~\cite{gpusyncwithscopes, gpucachecoherence, relativistic_cc, tc_coherence, acceleratingatomics, aim}. These approaches however, require the implementation of costly cache coherence protocols for GPUs.

\textbf{Leveraging cache coherence protocols for atomics processing. }
Prior works for CPUs~\cite{phi, rich, coup, ccache} add hardware close to caches to enable processing of atomic commutative operations, and modify the cache coherence protocol to aggregate commutative atomic operations across cores of the multiprocessor. Prior work that ain to accelerate atomics in GPUs propose change to cache coherence protocols to handle atomic requests GPU~\cite{gpucachecoherence, relativistic_cc, tc_coherence, acceleratingatomics, hlrc}.
However, these works require non-trivial changes to GPUs cache coherence protocols at the L1. Additionally, similar to the L1 buffering approach, it does not solve the contention in the LSU units when there are a large number of atomic updates.

\textbf{Software approaches for warp-level reduction.}
Software frameworks~\cite{cccl, cub, warpredblog, warpred_dsl, massiveatomics} and libraries provide functions that perform warp-level and block-level reduction. Using these frameworks results in a slowdown since the function has to be called for every atomic update, on a dynamically determined number of active threads producing the atomic updates. We compare with the CCCL library in~\cref{sec:results_cccl} and demonstrate that using it for differentiable rendering workloads leads to a slowdown. With \SWprojname, we propose efficient implementations that perform updates to all parameters associated with a primitive with a single function call.

\section{Conclusion}
We introduce \projname, a novel primitive that enables fast processing of atomic reduction operations in applications that \textbf{(1)} generate a massive number of atomic requests, and \textbf{(2)} have many threads within each warp atomically updating a common parameter. 
The key ideas behind \projname are to perform some atomic aggregation using warp-level reduction in SM sub-cores and distribute the atomic operations between the core and the L2 atomic units to efficiently utilize both. 
We implement an open-source software-only version of \projname.
We demonstrate that \projname can effectively alleviate the atomic processing bottleneck to accelerate raster-based differentiable rendering workloads, an important emerging class of applications in visual computing.



\bibliographystyle{IEEEtran}
\bibliography{refs}

\end{document}